\documentclass[final]{cvpr}

\usepackage{times}
\usepackage{epsfig}
\usepackage{graphicx}
\usepackage{amsmath}
\usepackage{amssymb}
\usepackage{multirow}
\usepackage{gensymb}
\usepackage{booktabs}
\usepackage{tabulary}
\usepackage[dvipsnames]{xcolor}
\usepackage[caption=false]{subfig}
\usepackage[british,english,american]{babel}
\usepackage{soul}
\usepackage{xspace}\xspace
\usepackage{adjustbox}
\usepackage{array}

\usepackage{dblfloatfix}
\usepackage{transparent}

\usepackage{algorithm}
\usepackage{listings}

\usepackage{etoolbox}
\makeatletter
\AfterEndEnvironment{algorithm}{\let\@algcomment\relax}
\AtEndEnvironment{algorithm}{\kern2pt\hrule\relax\vskip3pt\@algcomment}
\let\@algcomment\relax
\newcommand\algcomment[1]{\def\@algcomment{\footnotesize#1}}
\renewcommand\fs@ruled{\def\@fs@cfont{\bfseries}\let\@fs@capt\floatc@ruled
  \def\@fs@pre{\hrule height.8pt depth0pt \kern2pt}%
  \def\@fs@post{}%
  \def\@fs@mid{\kern2pt\hrule\kern2pt}%
  \let\@fs@iftopcapt\iftrue}
\makeatother

\definecolor{citecolor}{HTML}{0071bc}
\usepackage[pagebackref=true,breaklinks=true,colorlinks,citecolor=citecolor,bookmarks=false]{hyperref}

\newcommand{\app}{\raise.17ex\hbox{$\scriptstyle\sim$}}

\newcolumntype{x}[1]{>{\centering\arraybackslash}p{#1pt}}
\newcolumntype{y}[1]{>{\raggedright\arraybackslash}p{#1pt}}
\newcolumntype{z}[1]{>{\raggedleft\arraybackslash}p{#1pt}}
\newlength\savewidth\newcommand\shline{\noalign{\global\savewidth\arrayrulewidth
  \global\arrayrulewidth 1pt}\hline\noalign{\global\arrayrulewidth\savewidth}}
\newcommand{\tablestyle}[2]{\setlength{\tabcolsep}{#1}\renewcommand{\arraystretch}{#2}\centering\footnotesize}

\makeatletter\renewcommand\paragraph{\@startsection{paragraph}{4}{\z@}
  {.5em \@plus1ex \@minus.2ex}{-.5em}{\normalfont\normalsize\bfseries}}\makeatother

\definecolor{Gray}{gray}{0.5}

\newcommand{\cgap}[2]{
\fontsize{7pt}{1em}\selectfont{(${#1}${#2})}
}

\newcommand{\demph}[1]{\textcolor[rgb]{0.51,0.51,0.51}{#1}}
\newcommand{\cgreen}[2]{{#1}\textcolor[rgb]{0.37,0.69,0.34}{\cgap{+}{\textbf{#2}}}}	
\newcommand{\bfcgreen}[2]{{\textbf{#1}}\textcolor[rgb]{0.37,0.69,0.34}{\cgap{+}{\textbf{#2}}}}	
\newcommand{\cgray}[2]{{#1}\textcolor[rgb]{0.51,0.51,0.51}{\cgap{-}{\textbf{#2}}}}	

\makeatletter\renewcommand\paragraph{\@startsection{paragraph}{4}{\z@}
  {.5em \@plus1ex \@minus.2ex}{-.5em}{\normalfont\normalsize\bfseries}}\makeatother

\def\cvprPaperID{2358} 
\def\confYear{CVPR 2021}

\begin{document}

\title{Quality-Aware Network for Human Parsing}

\author{Lu Yang$^1$, Qing Song\thanks{The corresponding author is Qing Song.} $^1$, Zhihui Wang$^1$, Zhiwei Liu$^2$, Songcen Xu$^3$ and Zhihao Li$^3$\\
$^1$Beijing University of Posts and Telecommunications\\
$^2$Institute of Automation Chinese Academy of Sciences \\
$^3$Noah's Ark Lab, Huawei Technologies \\
{\tt\small \{soeaver, priv, wangzh\}@bupt.edu.cn} \\
{\tt\small zhiwei.liu@nlpr.ia.ac.cn, \{xusongcen, zhihao.li\}@huawei.com}
}

\maketitle

\begin{abstract}
How to estimate the quality of the network output is an important issue, and currently there is no effective solution in the field of human parsing. In order to solve this problem, this work proposes a statistical method based on the output probability map to calculate the pixel quality information, which is called pixel score. In addition, the Quality-Aware Module (QAM) is proposed to fuse the different quality information, the purpose of which is to estimate the quality of human parsing results. We combine QAM with a concise and effective network design to propose Quality-Aware Network (QANet) for human parsing. Benefiting from the superiority of QAM and QANet, we achieve the best performance on three multiple and one single human parsing benchmarks, including CIHP, MHP-v2, Pascal-Person-Part and LIP. Without increasing the training and inference time, QAM improves the AP$^\text{r}$ criterion by more than 10 points in the multiple human parsing task. QAM can be extended to other tasks with good quality estimation, \eg instance segmentation. Specifically, QAM improves Mask R-CNN by \app1\% mAP on COCO and LVISv1.0 datasets. Based on the proposed QAM and QANet, our overall system wins 1st place in CVPR2019 COCO DensePose Challenge, and 1st place in Track 1 \& 2 of CVPR2020 LIP Challenge. Code and models are available at \url{https://github.com/soeaver/QANet}.

\end{abstract}

\section{Introduction}

Predicting the semantic category of each pixel on the human body is a fundamental task in multimedia and computer vision, which usually known as human parsing~\cite{Liang_tpami2018_lip, Zhao_mm2018_mhpv2, Gong_eccv2018_pgn, Yang_cvpr2019_parsingrcnn, Wang_cvpr2020_hhp}. According to the number of human in the image, it can be divided into single human parsing~\cite{Liang_tpami2018_lip, Wang_iccv2019_cnif, Wang_cvpr2020_hhp} and multiple human parsing~\cite{Gong_eccv2018_pgn, Yang_cvpr2019_parsingrcnn, Yang_eccv2020_rprcnn}. Due to the diverse human appearance, semantic ambiguity of different human parts, and complex background, the practical application of human parsing is still difficult.

\begin{figure}
\begin{center}
\includegraphics[width=0.98\linewidth]{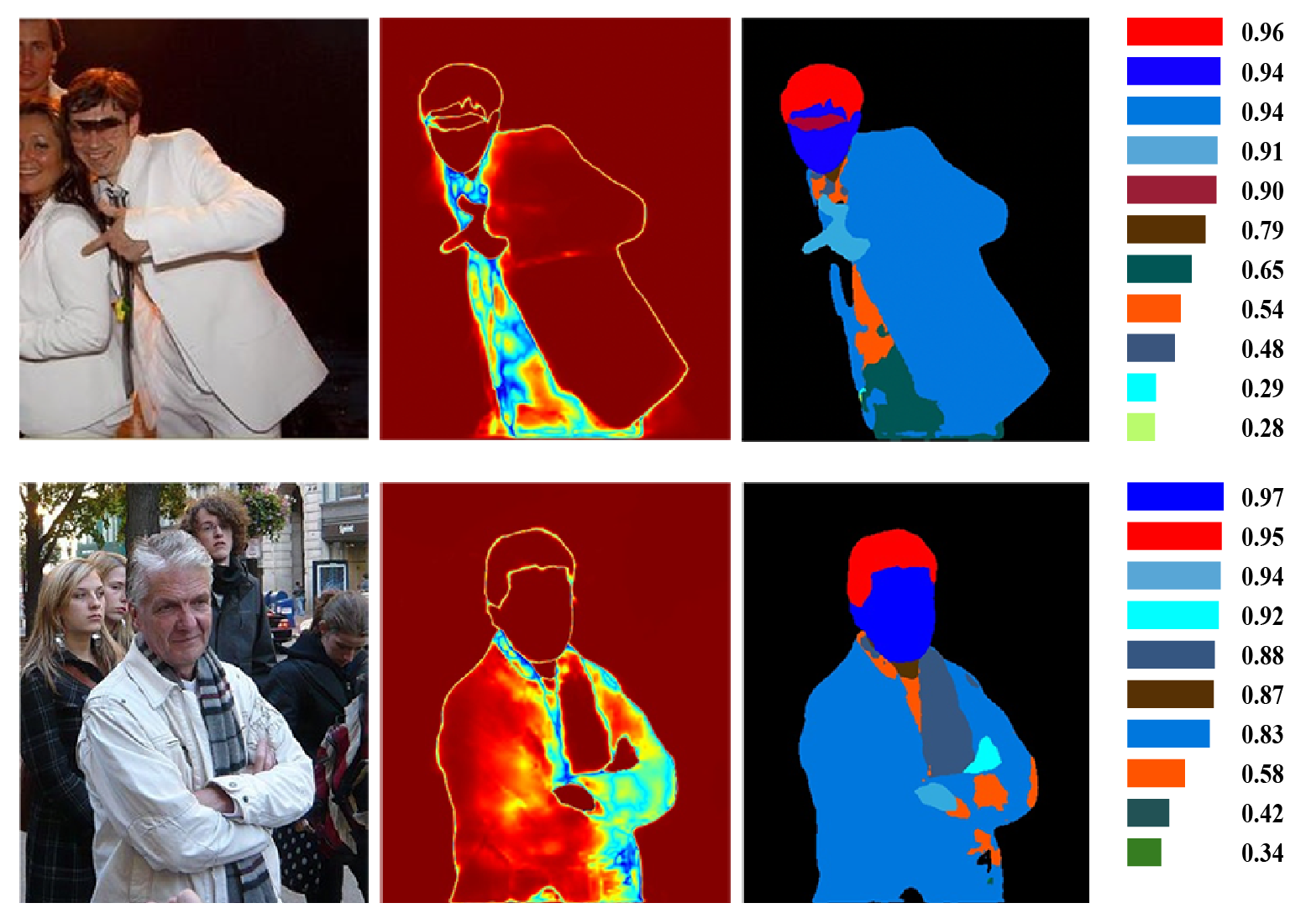}
\end{center}
\vspace{-1mm}
\caption{\textbf{Examples of category pixel score on CIHP \texttt{val} set}. Category pixel score can well reflect the quality of parsing results in different human parts.}
\label{fig:cate_pixel_score}
\vspace{-.8em}
\end{figure}

In recent years, many efforts have been made to improve the performance of human parsing. Structured hierarchy~\cite{Wang_iccv2019_cnif, Wang_cvpr2020_hhp, Ji_eccv2020_sematree}, edge and pose supervisions~\cite{Nie_eccv2018_mutual, Ruan_aaai2019_ce2p, Zhang_cvpr2020_corrpm}, attention mechanism~\cite{Yang_cvpr2019_parsingrcnn, Yang_eccv2020_rprcnn, Yuan_eccv2020_ocr} and self-correction~\cite{Li_arxiv2019_schp} have made remarkable progress. However, one important aspect has not been paid enough attention to, the quality estimation of human parsing results. The purpose of quality estimation is to score the prediction results of the system, thereby ranking outputs~\cite{Cao_icml2007_learning, Liang_acmsigir2020_setrank}, filtering out low-quality results~\cite{Philipp_cvpr2020_serfiq} or eliminating data noise~\cite{Li_arxiv2019_schp, Chang_cvpr2020_data}. Quality estimation has great potential in performance analysis and practical application for human parsing.  There are a large number of network outputs that have a low-score with high-quality, or a high-score with low-quality. In practical application, the parsing scores of most methods can only reflect the quality of detection results, but can not reflect the quality of parsing results~\cite{Zhao_mm2018_mhpv2, Yang_cvpr2019_parsingrcnn, Ruan_aaai2019_ce2p, Qin_bmvc2019_unified}, which makes it difficult to effectively filter out low-quality results by threshold. Moreover, in some researches, parsing results will be output in the form of a structured hierarchy~\cite{Gong_cvpr2019_graphonomy, Wang_iccv2019_cnif, Wang_cvpr2020_hhp, Ji_eccv2020_sematree}. In these methods, the score of each human part is the same and does not reflect the quality of semantic segmentation, so it is difficult to select the optimal level as the output. 

The purpose of this work is to accurately estimate the quality of human parsing results. We use the box score predicted by the detector as the \textit{\textbf{discriminant quality information}}. We extract the high confidence region from the probability map of network output, and take the average confidence of this region as the \textit{\textbf{pixel quality information}}, which is recorded as the pixel score (Figure~\ref{fig:cate_pixel_score}). Referring ~\cite{Huang_cvpr2019_msrcnn} and ~\cite{Yang_eccv2020_rprcnn}, we also construct a lighter FCN to regress the IoU between the network output and the ground-truth, which is used to express the \textit{\textbf{task quality information}} of the network output and record it as the IoU score. On this basis,  we present {\emph{Quality-Aware Module}} (QAM) as a way of building comprehensive and concise quality estimation for human parsing. QAM fuses the different quality information to generate a quality score to express the quality of human parsing result by using a flexible exponential weighting way. QAM is a post-processing mechanism independent of network structure, which can be combined with various methods. In this paper, we follow the simple and effective network design concept~\cite{Xiao_eccv2018_simple}, combined with QAM to propose a network suitable for both single and multiple human parsing, called {\emph{Quality-Aware Network}} (QANet). QANet takes ResNet-FPN~\cite{He_cvpr2016_resnet, Lin_cvpr2017_fpn} or HRNet~\cite{Sun_cvpr2019_hrnet} as the backbone, generates high-resolution semantic features through semantic FPN~\cite{Kirillov_cvpr2019_pfpn}, and predicts parsing result and IoU score by two different branches. Finally, QANet uses the quality score to express the quality of human parsing results.

We extensively evaluate our approach on three multiple and one single human parsing benchmarks, including CIHP~\cite{Gong_eccv2018_pgn}, MHP-v2~\cite{Zhao_mm2018_mhpv2}, Pascal-Person-Part~\cite{Xia_cvpr2017_ppp} and LIP~\cite{Liang_tpami2018_lip}. QANet has achieved state-of-the-art performances on these four benchmarks. Especially in terms of AP$^\text{r}$ criterion, QANet is ahead of ~\cite{Yang_cvpr2019_parsingrcnn, Ruan_aaai2019_ce2p, Yang_eccv2020_rprcnn, Ji_eccv2020_sematree} by more than 10 points on CIHP \texttt{val} set. In addition, QAM is a post-processing quality estimation mechanism, which can be combined with other advanced methods. Experiments show that QAM can be directly integrated into the region-based multiple human parsing methods such as Parsing R-CNN~\cite{Yang_cvpr2019_parsingrcnn} and RP R-CNN~\cite{Yang_eccv2020_rprcnn}, and can significantly improve the performance of both methods. QAM also has good quality estimation ability in instance segmentation task, which can improve Mask R-CNN by \app1\% mAP on COCO~\cite{Lin_eccv2014_coco} and LVISv1.0~\cite{Gupta_cvpr2019_lvis} datasets. The performance improvement effect of QAM is equivalent to ~\cite{Huang_cvpr2019_msrcnn}, but QAM does not require any additional training and inference time. These results show that QAM is a simple, modularized, low-cost and easily extensible module. We have reason to believe that it can be used in more tasks~\cite{Wei_cvpr2016_cpm, Kirillov_cvpr2019_ps, Georgia_iccv2019_meshrcnn, Yang_tip2020_hier}. 

Benefiting from the superiority of QANet, our overall system wins 1st place in CVPR2019 COCO DensePose Challenge\footnote{\fontsize{7pt}{1em}\url{https://cocodataset.org/}}, and 1st place in Track 1 \& 2 in CVPR2020 LIP Challenge\footnote{\fontsize{7pt}{1em}\url{https://vuhcs.github.io/}}.

\section{Related Work}
\label{sec:rel}

\noindent\textbf{Human Parsing.} Thanks to the progress of convolutional neural network~\cite{Alex_nip2012_alexnet, He_cvpr2016_resnet, Ren_nips2015_faster-rcnn, Long_cvpr2015_fcn} and open sourcing of large-scale human parsing datasets~\cite{Liang_tpami2018_lip, Zhao_mm2018_mhpv2, Gong_eccv2018_pgn}, human parsing has made great progress. For single human parsing, most of the work is devoted to introducing human body structure prior and improving network design. Wang \etal~\cite{Wang_iccv2019_cnif, Wang_cvpr2020_hhp} use deep graph networks and hierarchical human structures to exploit the human representational capacity. ~\cite{Ruan_aaai2019_ce2p, Zhang_cvpr2020_corrpm} enhance the semantic information of human body by introducing edge or pose supervisions. Li \etal~\cite{Li_arxiv2019_schp} introduce a purification strategy to tackle the problem of learning with label noises, to promote the reliability of the supervised labels as well as the learned models. In the aspect of multiple human parsing, many effective top-down frameworks have been proposed. According to whether the human detector~\cite{Ren_nips2015_faster-rcnn, Tian_iccv2019_fcos, Zhu_wacv2021_cpm} and human parser are end-to-end trained, the top-down methods can be divided into one-stage and two-stage. One-stage top-down is a multi task learning system based on R-CNN~\cite{Girshick_cvpr2014_rcnn}. Yang \etal present Parsing R-CNN~\cite{Yang_cvpr2019_parsingrcnn} and RP R-CNN~\cite{Yang_eccv2020_rprcnn}, which introduce multi-scale features~\cite{Chen_tpami2016_deeplab, Chen_eccv2018_deeplabv3plus}, attention mechanism~\cite{Wang_cvpr2018_nonlocal, Yang_tnnls2018_airnet} and global semantic information to enhance the human visual perception ability of convolutional neural network, making breakthroughs in multiple human parsing and dense pose estimation~\cite{Guler_cvpr2018_densepose} tasks. Two-stage top-down is a more flexible solution and the best-performing currently. Generally speaking, a single human parsing network can be used as the parser for two-stage top-down method~\cite{Ruan_aaai2019_ce2p, Li_arxiv2019_schp}.

\vspace{6pt}
\noindent\textbf{Quality Estimation.} The purpose of quality estimation (or quality assessment) is to score or rank the prediction results of a system. The cross-entropy loss function can be regarded as an intuitive quality estimation, and most discriminant models use it to express the confidence of their results, such as image classification~\cite{Alex_nip2012_alexnet, He_cvpr2016_resnet}, object detection~\cite{Ren_nips2015_faster-rcnn, Liu_eccv2016_ssd} and semantic segmentation~\cite{Long_cvpr2015_fcn}. However, more and more systems need customized quality estimation methods which are closer to their problems to accurately score or rank the outputs. In the research of recommendation system, ~\cite{Liang_acmsigir2020_setrank, Pei_acmrs2019_rerank} use the list-wise loss function to compare the relationship between objects to achieve the ranking of the results. In order to build a high performance face recognition system, ~\cite{Philipp_cvpr2020_serfiq} proposes an unsupervised estimation method to filter out  low-quality face images. Huang \etal~\cite{Huang_cvpr2019_msrcnn} use a mask IoU head to learn the quality of the predicted instance masks. Similarly, Yang \etal~\cite{Yang_eccv2020_rprcnn} present a parsing re-scoring network to sense the quality of human parsing maps and gives appropriate scores.

However, for human parsing, the existing methods can not accurately express the quality of results, and can not effectively estimate the quality of each human part independently.

\section{Methodology}
\label{sec:method}

\subsection{Motivation}

The quality estimation of human parsing is an important but long neglected issue. Inappropriate scoring or ranking the network outputs results in  a large number of low-score with high-quality or high-score with low-quality outputs. Meanwhile, it is difficult for us to filter out low-quality results by threshold. Therefore, our goal is to propose a simple, low-cost solution, which can effectively score the results of human parsing, and can also give the independent score of each human part.

\subsection{Estimating the Quality with Probability Map}

\begin{figure}
\begin{center}
\includegraphics[width=0.95\linewidth]{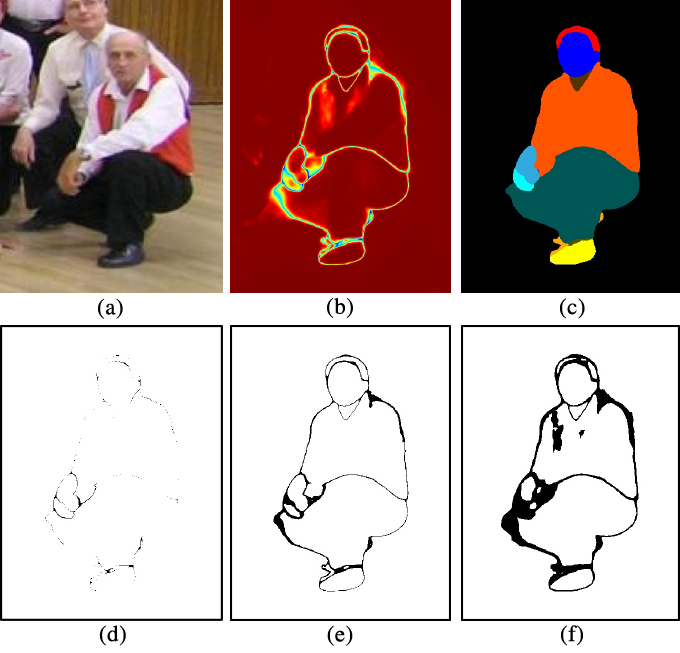}
\end{center}
\vspace{-1mm} 
\caption{\textbf{Probability map and high confidence masks with different thresholds}. (a) input image, (b) probability map, (c) parsing result, (d) high confidence mask with $T=0.4$, (e) $T=0.6$, (f) $T=0.8$. For (d) - (e), confidences of the black positions are lower than the threshold $T$.}
\label{fig:mask}
\vspace{-.8em}
\end{figure}

Formally, the output of the human parsing network is a feature map $\textit{\textbf{f}}\!\in\!\mathbb{R}^{N\!\times\! C\!\times\! H\!\times\! W\!}$, where $N$ denotes the number of human instances, $C$ denotes the number of parsing categories, $H$ and $W$ are height and width of feature maps. For the feature map \textit{\textbf{f}}, we take the largest probability prediction among the $C$ categories for each spatial position, and record it as probability map $\textit{\textbf{p}}\!\in\!\mathbb{R}^{N\!\times\! H\!\times\! W\!}$. Each position in \textit{\textbf{p}} expresses the classification confidence of corresponding pixel. Therefore, the probability map can be regarded as the quality estimation of each pixel. By making full use of the pixel quality information of the probability map, we can estimate the quality of the whole human body or part. A na\"ive solution is to average the confidence of all positions in the probability map to represent the quality of the parsing result. However, this scheme is not very accurate. Due to the mechanism of neural network is still difficult to explain, the output of some difficult samples is always full of randomness~\cite{Zhou_tpami2018_interp, Zhou_eccv2018_interp}. As shown in Figure~\ref{fig:mask}, some low confidence regions on the probability map often correspond to the boundary of the instance and some easily confused areas, but this does not mean that the prediction is wrong. For example, for a human instance with many boundaries or a human part with small area, the average of the probability map will be lower no matter whether the prediction are accurate or not. Therefore, if these areas are included in the average calculation, it will bring serious deviation to the quality estimation. 

\begin{algorithm}[t]
\caption{Pseudocode of Pixel Score in a PyTorch-like style.}
\label{alg:code}
\algcomment{\fontsize{7.2pt}{0em}\selectfont
}
\definecolor{codeblue}{rgb}{0.25,0.5,0.5}
\lstset{
  backgroundcolor=\color{white},
  basicstyle=\fontsize{7.2pt}{7.2pt}\ttfamily\selectfont,
  columns=fullflexible,
  breaklines=true,
  captionpos=b,
  commentstyle=\fontsize{7.2pt}{7.2pt}\color{codeblue},
  keywordstyle=\fontsize{7.2pt}{7.2pt},
}
\begin{lstlisting}[language=python]
# probs: input probability map (N, C, H, W)
# T: threshold, default 0.2
# C: num_classes
# inst_ps: instance pixel score
# cate_ps: category pixel score

# predicted category and probability	
inst_max, inst_idx = max(probs, dim=1)

# high confidence mask  (hcm)
inst_hcm = (inst_max >= T).to(bool)
# high confidence value  (hcv)
inst_hcv = sum(inst_max * inst_hcm, dim=[1, 2])
# number of hcm
inst_hcm_num = sum(inst_hcm, dim=[1, 2])

# instance pixel score, (N, 1)
inst_ps = inst_hcv / inst_hcm_num

cate_ps = one((N, C - 1))
for c in range(1, C):
    # probability of each predicted category 
    cate_max = inst_max * (inst_idx == c).to(bool)
    
    cate_hcm = (cate_max >= T).to(bool)
    cate_hcv = sum(cate_max * cate_hcm, dim=[1, 2])
    cate_hcm_num = sum(cate_hcm, dim=[1, 2])
    
    # category pixel score, (N, C)
    cate_ps[:, c - 1] = cate_hcv / cate_hcm_num

return inst_ps, cate_ps
\end{lstlisting}
\end{algorithm}

\begin{figure*}
\begin{center}
\includegraphics[width=0.98\linewidth]{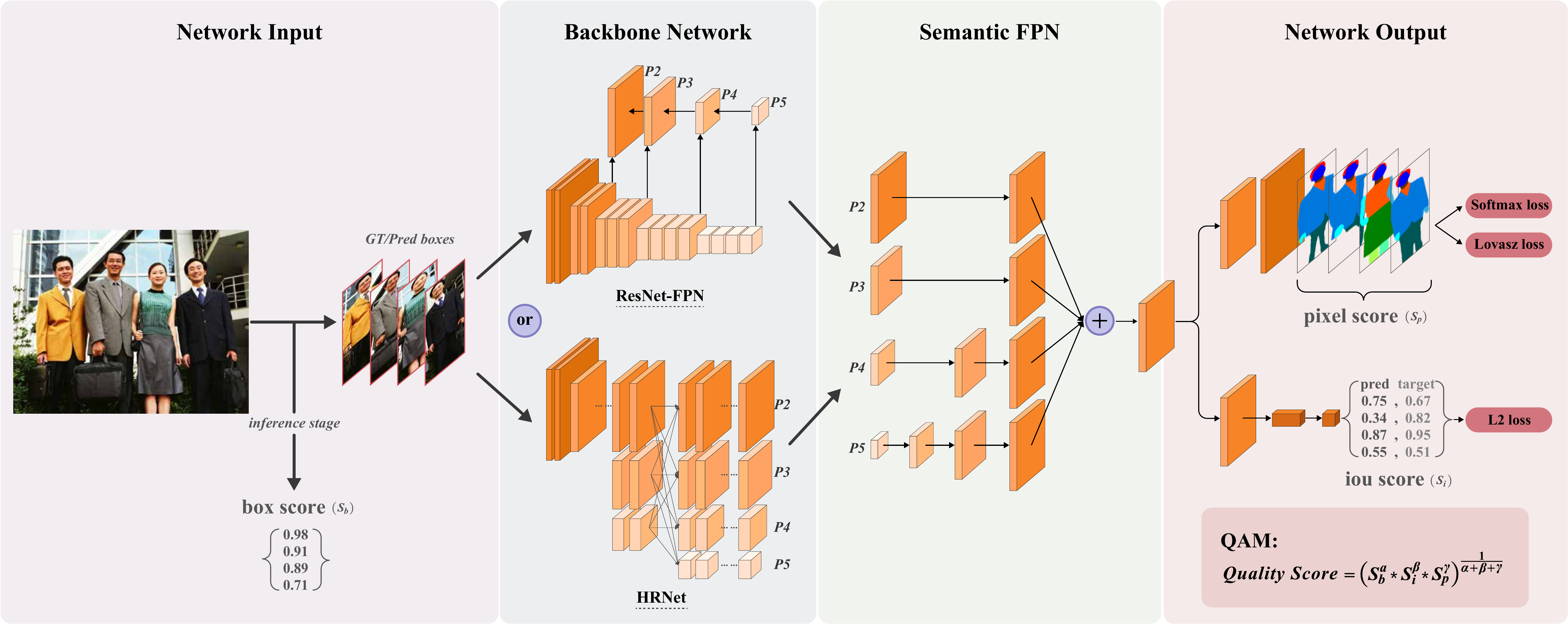}
\end{center}
\vspace{-1mm}
\caption{\textbf{QANet pipeline}. The whole image is cropped into several human instance according to the ground-truth boxes (during training) or predicted boxes (during inference), which are sent to the backbone network. The backbone network outputs multi-scale features through ResNet-FPN or HRNet. Semantic FPN integrates multi-scale features into high-resolution feature, and then predicts parsing result and its IoU score by two branches. During inference, we use QAM to estimate the quality of the predicted results for each human instance and each human part.}
\label{fig:qanet}
\vspace{-.8em}
\end{figure*}

To solve the above issue, we adopt a simple and effective scheme. We only calculate the average value of regions with confidence greater than threshold $T$, and the default $T=0.2$. More specifically, we calculate the {\emph{high confidence mask}} ($\textit{\textbf{hcm}}\!\in\!\mathbb{R}^{H\!\times\! W\!}$) on the probability map according to the threshold $T$. Then the average confidence of \textit{\textbf{hcm}} is calculated to express the pixel quality information of the human parsing network output, which is recorded as the instance pixel score. In addition, we also use the same method to calculate the pixel information quality of each predicted human part, which is recorded as category pixel score. Algorithm~\ref{alg:code} provides the PyTorch-like style pseudo-code of calculating instance pixel score and category pixel score. This calculation process is model independent and has less computational overhead.

\subsection{Quality Aware Module}

In addition to the pixel quality information of the probability map, some other information can be used to estimate the quality of human parsing results. Detector not only outputs the location information of human instance, but also gives the confidence of classification, which is called discriminant quality information. {\emph{Mean intersection over union}} (mIoU)~\cite{Long_cvpr2015_fcn} is the basic index to measure the parsing performance. Referring ~\cite{Huang_cvpr2019_msrcnn} and ~\cite{Yang_eccv2020_rprcnn}, we can regress the mIoU between the predicted parsing result and the ground-truth through a lightweight FCN, which represents the task quality information. Discriminant quality information, task quality information and pixel quality information all express the output quality of convolution neural network in some aspects, and they are complementary to each other.

We propose {\emph{Quality-Aware Module}} (QAM), which fuses the different quality information to estimate the output quality. In human parsing task, the box score $\textbf{\textit{S}}_{b}$ represents discriminant quality information, IoU score $\textbf{\textit{S}}_{i}$ represents task quality information and pixel score $\textbf{\textit{S}}_{p}$ represents pixel quality information. We use the exponential weighting to calculate the final quality score $\textbf{\textit{S}}_{q}$:
\begin{equation}
\begin{aligned}
\textbf{\textit{S}}_{q}\!=\!(\textbf{\textit{S}}_{b}^\alpha * \textbf{\textit{S}}_{i}^\beta * \textbf{\textit{S}}_{p}^\gamma)^\frac{1}{\alpha + \beta + \gamma},
\end{aligned}
\label{eq:quality_score}
\vspace{-1pt}
\end{equation}
where $\alpha$, $\beta$ and $\gamma$ are used to adjust the weight of different quality information, $\alpha=\beta=\gamma=1.0$ by default. When using the default weights, QAM can significantly improve the effect of human parsing quality estimation, and can get better results by adjusting the weights. The idea of QAM is to flexibly fuse the different quality information to generate a more comprehensive quality score, which is a simple, low-cost and easily extensible method. QAM can be used not only for human parsing, but also for instance segmentation~\cite{He_iccv2017_maskrcnn, Wang_nips2020_solov2} and dense pose estimation~\cite{Yang_cvpr2019_parsingrcnn} after simple adaptation.

\subsection{Architecture}

In order to verify the effect of QAM, we propose a two-stage top-down human parsing network, called {\emph{Quality-Aware Network}} (QANet). Our design principle starts from SimpleNet~\cite{Xiao_eccv2018_simple}, a strong and concise pose estimation baseline. As illustrated in Figure~\ref{fig:qanet}, we adopt a ResNet~\cite{He_cvpr2016_resnet} with FPN~\cite{Lin_cvpr2017_fpn} or HRNet~\cite{Sun_cvpr2019_hrnet} as backbone, which produces multi-scale features $P$2 - $P$5. Each scale feature is upsampled by convolutions and bilinear upsampling until it reaches 1/4 scale, these outputs are integrated into a high-resolution feature by element-wise summed~\cite{Kirillov_cvpr2019_pfpn}. Two branches are attached to the high-resolution feature to predict the parsing results and IoU scores respectively. The parsing loss is a combination of the cross-entropy loss (${\cal{L}}_{\text{c}}$) and the lovasz loss (${\cal{L}}_{\text{z}}$) ~\cite{Berman_iccv2018_lovasz}. The IoU branch is a lightweight FCN, and uses MSE loss (${\cal{L}}_{\text{m}}$) to regress the mIoU between the predicted parsing result and the ground-truth. Thus, the loss function of the whole network is:
 \begin{equation}
\begin{aligned}
{\cal{L}} = {\cal{L}}_{\text{c}} + {\cal{L}}_{\text{z}} + {\cal{L}}_{\text{m}}.
\end{aligned}
\label{eq:total_loss}
\vspace{-1pt}
\end{equation}
Finally, we use QAM to calculate the quality scores of the whole human instance and each human part respectively.

\section{Experiments}
\label{sec:exp}

\subsection{Experimental Settings} 

\noindent\textbf{Datasets.} Three multiple~\cite{Gong_eccv2018_pgn,  Zhao_mm2018_mhpv2, Xia_cvpr2017_ppp} and one single~\cite{Liang_tpami2018_lip} human parsing benchmarks are used for performance evaluation. {\emph{CIHP}}~\cite{Gong_eccv2018_pgn} is a popular multiple human parsing dataset that has 28,280 images for training, 5,000 images for validation and 5,000 images for testing with 20 categories. {\emph{MHP-v2}}~\cite{Zhao_mm2018_mhpv2} includes 25,403 elaborately annotated images with 58 fine-grained semantic category labels. The validation set and test set have 5,000 images respectively. The rest 15,403 are provided as the training set. {\emph{PASCAL-Person-Part}}~\cite{Xia_cvpr2017_ppp} contains 3,535 annotated images distributed to 1,717 for training and 1,818 for testing, among which only 7 categories are labeled. {\emph{LIP}}~\cite{Liang_tpami2018_lip} is a single human parsing dataset and contains 50,462 images, which are collected from realistic scenarios and divided into 30,462 images for training, 10,000 for validation and 10,000 for testing. 

\vspace{6pt}
\noindent\textbf{Evaluation Metrics.} The standard mIoU criterion is adopted for evaluation on human part segmentation, both for single\footnote{\fontsize{7pt}{1em}The quality estimation can not be reflected in the mIoU criterion, so we use QANet without QAM for single human parsing.} and multiple human parsing. We also use {\emph{Average Precision based on part}} (AP$^\text{p}$), {\emph{Average Precision based on region}} (AP$^\text{r}$) and {\emph{Percentage of Correctly parsed semantic Parts}} (PCP$_\text{50}$) to measure the performance of multiple human parsing, following~\cite{Gong_eccv2018_pgn, Zhao_mm2018_mhpv2, Yang_cvpr2019_parsingrcnn}.

\vspace{6pt}
\noindent\textbf{Training.} We implement the QAM and QANet based on Pytorch on a server with 8 NVIDIA Titan RTX GPUs. We use ResNet50, ResNet101 and HRNet-W48 as the backbones to compare different methods fairly. In particular, we only conduct HRNet-W48 experiments on LIP. Following ~\cite{Xiao_eccv2018_simple}, the ground truth human box is made to a fixed aspect ratio (4 : 3) by extending the box in height or width. The default input size of QANet is 512$\times$384. Data augmentation includes scale [-30\%, +30\%], rotation [-40\degree, +40\degree] and horizontal flip. For optimization, we use the ADAM~\cite{Diederik_iclr2015_adam} solver with 256 batch size. All models are trained for 140 epochs with 2e-3 base learning rate, which drops to 2e-4 at 90 epochs and 2e-5 at 120 epochs. No other training techniques are used, such as warm-up~\cite{Goyal_arxiv2017_1hour}, syncBN~\cite{Chao_cvpr2018_megdet}, learning rate annealing~\cite{Chen_tpami2016_deeplab, Zhao_cvpr2017_pspnet}.

\vspace{6pt}
\noindent\textbf{Inference.} For multiple human parsing, a two-stage top-down paradigm is applied~\cite{Chen_cvpr2018_cpn, Xiao_eccv2018_simple}. By default we use a FCOS~\cite{Tian_iccv2019_fcos} with ResNet50 as detector. The QANet result is the combination of the original input and the flipped image. For evaluating on LIP, we also use multi-scale test augmentation, following ~\cite{Wang_iccv2019_cnif, Wang_cvpr2020_hhp}.

\subsection{Ablation Studies} 

In this sub-section, we assess the effects of different settings on QAM and QANet by details ablation studies. 

\vspace{6pt}
\noindent\textbf{Threshold of Pixel Score.} Threshold $T$ is an important hyper-parameter for calculating the pixel score (see Algorithm~\ref{alg:code}). The table below shows QANet-ResNet50 with different thresholds:
\begin{center}
\vspace{-.8em}
\tablestyle{2pt}{1.2}	
\scalebox{1.05}{
\begin{tabular}{c|x{32}x{32}x{32}x{32}x{32}}
threshold $T$                        & 0.0 & 0.2 & 0.4 & 0.6 & 0.8  \\
\shline
AP$^\text{p}$/AP$^\text{r}$ & 47.5/42.2 & \textbf{60.1/56.3} & 60.1/56.2 & 60.0/56.0 & 59.9/55.1 \\
\end{tabular}
}
\vspace{-.8em}
\end{center}
when $T = 0.0$, the effect of pixel score is negative. Increasing $T$ to filter out low confidence region can significantly improve the effect. Therefore, we choose $T = 0.2$. As illustrated in Figure~\ref{fig:cate_pixel_score}, (category) pixel score can well reflect the quality of parsing results in different human parts.

\begin{table}[t]
\centering
\small
\tabcolsep 0.04in 
\scalebox{0.92}{
\begin{tabular}{c|c|cc|cc}
& Quality Weights                         &     \multirow{2}{*}{AP$^\text{p}$}    &  \multirow{2}{*}{ AP$^\text{p}_\text{50}$}    &  \multirow{2}{*}{AP$^\text{r}$}  & \multirow{2}{*}{AP$^\text{r}_\text{50}$}  \\
& ($\alpha$, \quad $\beta$, \quad $\gamma$) &                                    &                                                                    &                                                  &                                                                 \\
\shline   
 (a) &                (1.0, \ \demph{0.0}, \ \demph{0.0})         &                        55.6                         &                                67.7                             &                      40.1                     &                            45.0                             \\ 
 (b) &                 (\demph{0.0}, \ 1.0, \ \demph{0.0})         &            \cgreen{56.9}{1.3}               &                    \cgreen{70.7}{3.0}                    &         \cgreen{44.3}{4.2}            &                \cgreen{50.0}{5.0}                    \\ 
 (c) &                 (\demph{0.0}, \ \demph{0.0}, \ 1.0)         &            \cgray{49.0}{6.6}                 &                    \cgray{60.8}{6.9}                      &         \cgreen{49.7}{9.6}            &                \cgreen{55.4}{10.4}                   \\ 
 (d) &                 (1.0, \ 1.0, \ \demph{0.0})         &            \bfcgreen{60.2}{4.6}            &                    \bfcgreen{74.5}{6.8}                 &         \cgreen{45.6}{5.5}            &                \cgreen{51.4}{6.4}                    \\  
 (e) &                 (1.0, \ \demph{0.0}, \ 1.0)         &            \cgreen{57.0}{1.4}               &                    \cgreen{70.2}{2.5}                    &         \cgreen{55.0}{14.9}          &                \cgreen{62.3}{17.3}                   \\ 
 (f) &                 (\demph{0.0}, \ 1.0, \ 1.0)         &            \cgreen{57.0}{1.4}               &                    \cgreen{70.9}{3.2}                    &         \cgreen{52.5}{12.4}          &                \cgreen{59.6}{14.6}                   \\ 
 (g) &                 (1.0, \ 1.0, \ 1.0)         &            \bfcgreen{60.2}{4.6}            &                    \bfcgreen{74.5}{6.8}                 &         \cgreen{54.0}{13.9}          &                \cgreen{61.2}{16.2}                    \\  
\hline   
 (h) &     \textbf{(1.0, \  0.5, \ 3.0)}        &            \cgreen{60.1}{4.5}               &                    \cgreen{74.3}{6.6}                    &         \bfcgreen{56.2}{16.1}       &                \bfcgreen{63.5}{18.5}                \\ 
\end{tabular}
}
\vspace{.2em}
  \caption{\textbf{Ablation study of Quality Weights}. All models are trained on CIHP \texttt{train} set and evaluated on CIHP \texttt{val} set.}
  \label{tab:ablation_qweights}
\vspace{-.8em}
\end{table}

\begin{table*}[t]
\centering
\small
\tabcolsep 0.04in 
\scalebox{0.96}{
\begin{tabular}{c|cc|c|ccc|cc}
Methods                               &   GT-box            &        QAM     &          mIoU             &    AP$^\text{p}$     & AP$^\text{p}_\text{50}$&  PCP$_\text{50}$  &     AP$^\text{r}$     &  AP$^\text{r}_\text{50}$   \\
\shline   
 \multirow{4}{*}{QANet-R50}  &                         &                     &           62.9              &          55.6             &                67.7               &            68.9            &           40.1             &              45.0                   \\
                                            & \checkmark        &                      & \bfcgreen{65.0}{2.1} & \cgray{53.6}{2.0}&    \cgray{61.9}{5.8}       & \bfcgreen{69.0}{0.1} & \cgray{38.0}{2.1}    & \cgray{42.8}{2.2}     \\ 
                                            &                            & \checkmark  & \cgreen{62.9}{0.0} & \cgreen{60.1}{4.5} &     \cgreen{74.3}{6.6}    & \cgreen{68.9}{0.0} & \cgreen{56.2}{16.1} & \cgreen{63.5}{18.5}   \\ 
                                            & \checkmark        & \checkmark   & \bfcgreen{65.0}{2.1} & \bfcgreen{62.4}{6.8} &\bfcgreen{76.3}{8.6}& \bfcgreen{69.0}{0.1} & \bfcgreen{58.3}{18.2} & \bfcgreen{66.0}{21.0}    \\ 
\end{tabular}
}
\vspace{.2em}
  \caption{\textbf{The effect of QAM with GT-box or predicted box}. All models are trained on CIHP \texttt{train} set and evaluated on CIHP \texttt{val} set.}
  \label{tab:ablation_gtbox}
\vspace{-.8em}
\end{table*}

\begin{table*}[t]
\centering
\small
\tabcolsep 0.04in 
\scalebox{0.96}{
\begin{tabular}{c|ccccc|c|ccc|cc}
Methods                               &   S-FPN              &     Lovasz     &      IoU      &   Large-Res    &        QAM     &          mIoU             &    AP$^\text{p}$     & AP$^\text{p}_\text{50}$&  PCP$_\text{50}$  &     AP$^\text{r}$     &  AP$^\text{r}_\text{50}$   \\
\shline   
 \multirow{5}{*}{SimpleNet}  &                           &                      &                     &                       &                      &           54.6              &          51.8             &                60.1               &            59.1            &           32.5             &              36.1                   \\
                                            \cline{7-12}
                                            & \checkmark        &                      &                     &                        &                      & \cgreen{56.2}{1.6} & \cgreen{53.6}{1.8} &    \cgreen{63.8}{3.7}     & \cgreen{62.0}{2.9} & \cgreen{34.7}{2.2} & \cgreen{38.4}{2.3}     \\ 
                                            & \checkmark        & \checkmark   &                     &                        &                      & \cgreen{58.9}{4.3} & \cgreen{53.9}{2.1} &    \cgreen{64.3}{3.2}     & \cgreen{65.9}{6.8} & \cgreen{37.2}{4.7} & \cgreen{41.9}{5.8}     \\ 
                                            & \checkmark        & \checkmark   & \checkmark  &                       &                      & \cgreen{59.3}{4.7} & \cgreen{58.3}{6.5} &     \cgreen{71.9}{11.8}   & \cgreen{66.0}{6.9} & \cgreen{41.8}{9.3} & \cgreen{47.1}{11.0}   \\ 
                                            & \checkmark        & \checkmark   & \checkmark  &  \checkmark   &                      & \bfcgreen{62.9}{8.3} &\bfcgreen{60.3}{8.5}&\bfcgreen{74.5}{14.4} & \bfcgreen{68.9}{9.8} & \cgreen{45.6}{13.1}& \cgreen{51.4}{15.3}    \\ 
                                            \hline
\textbf{QANet}                    & \checkmark        & \checkmark   & \checkmark  &  \checkmark    & \checkmark   & \bfcgreen{62.9}{8.3} &\cgreen{60.1}{8.3} &     \cgreen{74.3}{14.2} & \bfcgreen{68.9}{9.8} &\bfcgreen{56.2}{23.7}& \bfcgreen{63.5}{27.4} \\ 
\end{tabular}
}
\vspace{.2em}
  \caption{\textbf{Ablation study of QANet on CIHP dataset}. `S-FPN' is Semantic FPN, `Lovasz' is lovasz loss,  `IoU' denotes a lightweight FCN to predict IoU score, `Large-Res' denotes 512$\times$384 input resolution.}
  \label{tab:ablation_qanet}
\vspace{-.8em}
\end{table*}

\vspace{6pt}
\noindent\textbf{Quality Weights.} Quality weights is used to adjust the proportion of box score, IoU score and pixel score in quality estimation. Table~\ref{tab:ablation_qweights} shows the ablation study of quality weights on CIHP \texttt{val} set. All experiments are based on ResNet50. Row (a), (b) and (c) indicate the performance of human parsing when only box score, IoU score or pixel score is used. It can be seen that the IoU score improves both AP$^\text{p}$ and AP$^\text{r}$, but pixel improves AP$^\text{r}$ and degrades AP$^\text{p}$.  Row (d), (e) and (f) indicate any combination of two scores will improve the AP$^\text{p}$ and AP$^\text{r}$. Especially when the pixel score is used, the improvement of AP$^\text{r}$ is very significant. This proves pixel score is better at representing local (human part) quality. When box score, IoU score and pixel score are used with equal weights (Row (g)), AP$^\text{p}$ and AP$^\text{r}$ are increased by 4.6 and 13.9 points respectively. This fully proves that the pixel score and QAM proposed by this work are effective. After grid search, we use \textbf{(1.0, \  0.5, \ 3.0)} as the quality weights to get the best performance on CIHP. We also use this weight on both MHP-v2 and PASCAL-Person-Part. In addition, we use ground-truth boxes (GT-box) to replace the predicted boxes. The experiments (see Table~\ref{tab:ablation_gtbox}) shows that QAM can still significantly improve the performance of human parsing when using GT-box.

\vspace{6pt}
\noindent\textbf{From SimpleNet to QANet.} We adopt SimpleNet as our baseline, which uses ResNet50 as the backbone and followed by three deconvolutional layers~\cite{Xiao_eccv2018_simple}. By default, the input size of SimpleNet is 256$\times$192. As shown is Table~\ref{tab:ablation_qanet}, the SimpleNet achieves 54.6\% mIoU, 51.8\% AP$^\text{p}$ and 32.5\% AP$^\text{r}$ on CIHP \texttt{val} set. Semantic FPN, lovasz loss, IoU score and 512$\times$384 input size outperform the baseline by 8.3\% mIoU, 8.5\% AP$^\text{p}$ and 13.1\% AP$^\text{r}$. This performance has exceeded the best method on CIHP~\cite{Zhang_cvpr2020_corrpm, Yang_eccv2020_rprcnn, Ji_eccv2020_sematree}, which shows that our network architecture is concise but effective. Furthermore, our proposed QANet (with QAM) achieves 62.9\% mIoU, 60.1\% AP$^\text{p}$ and 56.2\% AP$^\text{r}$. QAM brings more than 10 points AP$^\text{r}$ improvement.

\subsection{Quantitative and Qualitative Results} 

\begin{table}[t]
\centering
\small
\tabcolsep 0.04in 
\scalebox{0.96}{
\begin{tabular}{l|c|ccc}
 Methods                                                                             & Input Size         &  pix Acc.            &  mean Acc.            &  mIoU   \\
  \shline  
FCN-8s~\cite{Long_cvpr2015_fcn}                                     & --                        & 76.06               & 36.75                    & 28.29    \\     
Attention~\cite{Chen_cvpr2016_attention-scale}                & --                        & 83.43               & 54.39                    & 42.92    \\             
MMAN~\cite{Luo_eccv2018_mman}                                  & 256$\times$256 & 85.24               & 57.60                    & 46.93    \\
JPPNet~\cite{Liang_tpami2018_lip}                                   & 384$\times$384 & --                      & --                          & 51.37    \\
CE2P~\cite{Ruan_aaai2019_ce2p}                                    & 473$\times$473 & 87.37               & 63.20                    & 53.10    \\
BraidNet~\cite{Liu_mm2019_braidnet}                               & 384$\times$384 & 87.60               & 66.09                    & 54.42    \\
CorrPM~\cite{Zhang_cvpr2020_corrpm}                            & 384$\times$384  & --                      & --                          & 55.33    \\
OCR~\cite{Yuan_eccv2020_ocr}                                        & 473$\times$473  & --                      & --                          & 56.65    \\
PCNet~\cite{Zhang_cvpr2020_pcnet}                                & 473$\times$473  & --                      & --                          & 57.03    \\
CNIF~\cite{Wang_iccv2019_cnif}                                       & 473$\times$473  & 88.03               & 68.80                    & 57.74    \\
HHP~\cite{Wang_cvpr2020_hhp}                                       & 473$\times$473  & \textbf{89.05}   & 70.58                   & 59.25    \\
SCHP~\cite{Li_arxiv2019_schp}                                         & 473$\times$473  & --                     & --                          & 59.36    \\
  \hline  
\textbf{QANet (ours)}                                                          & 512$\times$384  & 88.92               & \textbf{71.17}       & \textbf{59.61}    \\  
\end{tabular}
}
\vspace{.2em}
  \caption{\textbf{Comparison of pixel accuracy, mean accuracy and mIoU on LIP \texttt{val} set}. Our best single model achieves state-of-the-art with smaller input size ($512\times384 < 473\times473$).}
  \label{tab:lip}
\vspace{-.8em}
\end{table}

\begin{table*}[t]
\centering
\small
\tabcolsep 0.04in 
\scalebox{0.94}{
\begin{tabular}{c|l|l|c|c|ccc|cc}
               Datasets         &      Methods                                                                                         &    Backbones    &   \  Epochs   \quad &  \ mIoU \quad &  \quad AP$^\text{p}$ \quad  & \quad AP$^\text{p}_\text{50}$ \quad & \quad PCP$_\text{50}$ \quad & \quad AP$^\text{r}$ \quad & \quad  AP$^\text{r}_\text{50}$ \quad     \\
\shline  
 \multirow{17}{*}{CIHP~\cite{Gong_eccv2018_pgn}} & \multicolumn{2}{c|}{Bottom-Up} &\\
                                      \cline{2-10}       
                                      & PGN$^\dagger$~\cite{Gong_eccv2018_pgn}                                    & ResNet101      & \app80        & 55.8      & 39.0                  & 34.0                                & 61.0                      & 33.6                & 35.8                                  \\             
                                      & Graphonomy~\cite{Gong_cvpr2019_graphonomy}                            & Xception          & 100            & 58.6       & --                      & --                                     & --                           & --                      & --                                     \\
                                      & GPM~\cite{He_aaai2020_grapyml}                                                    & Xception          & 100            & 60.3       & --                      & --                                     & --                           & --                      & --                                     \\
                                      & Grapy-ML~\cite{He_aaai2020_grapyml}                                            & Xception          & 200            & 60.6       & --                      & --                                     & --                           & --                      & --                                     \\
                                      & CorrPM~\cite{Zhang_cvpr2020_corrpm}                                           & ResNet101       & 150            & 60.2      & --                      & --                                     & --                            & --                     & --                                     \\ 
                                      \cline{2-10}                                            
                                       & \multicolumn{2}{c|}{One-Stage Top-Down} &\\
                                      \cline{2-10}      
                                      & Parsing R-CNN~\cite{Yang_cvpr2019_parsingrcnn}                          & ResNet50        & 75              & 56.3      & 53.9                  & 63.7                                & 60.1                        & \demph{36.5}   & \demph{40.9}                    \\
                                      & Unified~\cite{Qin_bmvc2019_unified}                                                & ResNet101       & \app37       & 55.2      & 48.0                  & 51.0                                & --                            & 38.6                 & 44.0                                  \\
                                      & RP R-CNN~\cite{Yang_eccv2020_rprcnn}                                         & ResNet50         & 150            & 60.2     & 59.5                  & 74.1                                & 64.9                        & \demph{42.3}   & \demph{48.2}                    \\
                                      \cline{2-10}  
                                       & \multicolumn{2}{c|}{Two-Stage Top-Down} &\\
                                      \cline{2-10}       
                                      & CE2P~\cite{Ruan_aaai2019_ce2p}                                                   & ResNet101        & 150            & 59.5      & --                      & --                                     & --                           & 42.8                & 48.7                                 \\
                                      & BraidNet~\cite{Liu_mm2019_braidnet}                                              & ResNet101        & 150            & 60.6      & --                      & --                                     & --                           & 43.6                & 49.9                                 \\
                                      & SemaTree~\cite{Ji_eccv2020_sematree}                                          & ResNet101        & 200            & 60.9      & --                      & --                                     & --                           & 44.0                & 49.3                                  \\       
                                      & \textbf{QANet (ours)}                                                                         & ResNet50          & 140            & 62.9      & 60.1                 & 74.3                                 & 68.9                      & 56.2                & 63.5                                  \\       
                                      & \textbf{QANet (ours)}                                                                         & ResNet101        & 140            & 63.8      & 61.7                 & 77.1                                 & 72.0                      & 57.3                & 64.8                                  \\      
                                      & \textbf{QANet (ours)}                                                                         & HRNet-W48       & 140            & \textbf{66.1} &\textbf{64.5} & \textbf{81.3}                  & \textbf{75.7}          & \textbf{60.8}    & \textbf{68.8}                     \\       
\shline                                        
\multirow{12}{*}{MHP-v2~\cite{Zhao_mm2018_mhpv2}}& \multicolumn{2}{c|}{Bottom-Up} &\\
                                       \cline{2-10}      
                                      & MH-Parser~\cite{Li_arxiv2017_mhparser}                                         & ResNet101        & --                & --          & 36.0                  & 17.9                               & 26.9                      & --                      & --                                      \\
                                      & NAN~\cite{Zhao_mm2018_mhpv2}                                                   &  --                       & \app80       & --          & 41.7                  & 25.1                               & 32.2                      & --                      & --                                      \\
                                       \cline{2-10}                        
                                       & \multicolumn{2}{c|}{One-Stage Top-Down} &\\
                                       \cline{2-10}  
                                      & Parsing R-CNN~\cite{Yang_cvpr2019_parsingrcnn}                         & ResNet50           & 75             & 36.2      & 39.5                   & 24.5                               & 37.2                      & --                      & --                                      \\
                                      & RP R-CNN~\cite{Yang_eccv2020_rprcnn}                                        & ResNet50           & 150           & 38.6      & 46.8                   & 45.3                               & 43.8                      & --                      & --                                      \\
                                      \cline{2-10}                       
                                       & \multicolumn{2}{c|}{Two-Stage Top-Down} &\\
                                      \cline{2-10}       
                                      & CE2P~\cite{Ruan_aaai2019_ce2p}                                                 & ResNet101         & 150             & 41.1    & 42.7                    & 34.5                               & 43.8                      & --                      & --                                      \\
                                      & SemaTree~\cite{Ji_eccv2020_sematree}                                         & ResNet101         & 200             & --         & 42.5                   & 34.4                                & 43.5                      & --                      & --                                      \\
                                      & \textbf{QANet (ours)}                                                                         & ResNet50          & 140             & 42.4      & 47.2                 & 44.0                                 & 46.5                       & --                      & --                                     \\       
                                      & \textbf{QANet (ours)}                                                                         & ResNet101        & 140             & 43.1      & 49.2                 & 48.8                                 & 50.8                       & --                      & --                                     \\      
                                      & \textbf{QANet (ours)}                                                                         & HRNet-W48      & 140             & \textbf{44.4} &\textbf{51.0} & \textbf{54.0}                  & \textbf{55.8}           & --                      & --                                     \\         
\shline                                        
\multirow{13}{*}{PPP~\cite{Xia_cvpr2017_ppp}}& \multicolumn{2}{c|}{Bottom-Up} &\\                      
                                      \cline{2-10}       
                                      & PGN$^\dagger$~\cite{Gong_eccv2018_pgn}                                    & ResNet101      & $\sim$80    & 68.4       & --                      & --                                     & --                           & 39.2                & 39.6                                  \\             
                                      & Graphonomy~\cite{Gong_cvpr2019_graphonomy}                            & Xception          & 100            & 71.1       & --                      & --                                     & --                           & --                      & --                                     \\
                                      & GPM~\cite{He_aaai2020_grapyml}                                                    & Xception          & 100            & 69.5       & --                      & --                                     & --                           & --                      & --                                     \\
                                      & Grapy-ML~\cite{He_aaai2020_grapyml}                                            & Xception          & 200            & 71.6       & --                      & --                                     & --                           & --                      & --                                     \\
                                       \cline{2-10}                        
                                       & \multicolumn{2}{c|}{One-Stage Top-Down} &\\
                                       \cline{2-10}  
                                      & Parsing R-CNN~\cite{Yang_cvpr2019_parsingrcnn}                         & ResNet50           & 75             & \demph{62.7} & \demph{49.8} & \demph{58.2}          & \demph{48.7}        & \demph{40.4}   & \demph{43.7}                   \\
                                      & RP R-CNN~\cite{Yang_eccv2020_rprcnn}                                        & ResNet50           & 75           & \demph{63.3} & \demph{50.1} & \demph{58.9}          & \demph{49.1}        & \demph{40.9}   & \demph{44.1}                   \\
                                      \cline{2-10}  
                                       & \multicolumn{2}{c|}{Two-Stage Top-Down} &\\
                                      \cline{2-10}       
                                      & CNIF$^\dagger$~\cite{Wang_iccv2019_cnif}                                    & ResNet101        & 150            & 70.8      & --                      & --                                     & --                           & --                      & --                                    \\
                                      & HHP$^\dagger$~\cite{Wang_cvpr2020_hhp}                                    & ResNet101        & 150            & \textbf{73.1}      & --                      & --                                     & --                           & --                      & --                                    \\
                                      & \textbf{QANet (ours)}                                                                         & ResNet101        & 140            & 69.5      & 60.1                 & 74.6                                 & 62.9                       & 54.0                 & 62.6                                \\      
                                      & \textbf{QANet (ours)}$^\dagger$                                                       & HRNet-W48       & 140            & 72.6      &  \textbf{63.1}    & \textbf{78.2}                    & \textbf{67.2}           & \textbf{58.7}     & \textbf{67.8}                       \\       
\end{tabular}
}
\vspace{.2em}
  \caption{\textbf{Comparison with previous methods on multiple human parsing on CIHP, MHP-v2 and PASCAL-Person-Part (denoted as PPP) datasets}. \textbf{Bold} numbers are state-of-the-art on each dataset, and \demph{gray} numbers are based on the official open sourcing implementation by us~\cite{Yang_cvpr2019_parsingrcnn, Yang_eccv2020_rprcnn}. $^\dagger$ denotes using multi-scale test augmentation.}
  \label{tab:sota}
\vspace{-.8em}
\end{table*}

\noindent\textbf{LIP~\cite{Liang_tpami2018_lip}.} LIP is a standard benchmark for single human parsing. Table~\ref{tab:lip} shows QANet (HRNet-W48) with other state-of- the-arts on LIP \texttt{val} set. In order to ensure that the number of pixels of the input image is almost the same, we have carried out the experiments of 512$\times$384 input size, which is smaller than others. In terms of mean Acc. and mean IoU, our QANet surpasses the best performing method HHP$^\dagger$~\cite{Wang_cvpr2020_hhp} by 0.59 and 0.25 points, respectively.

\vspace{6pt}
\noindent\textbf{CIHP~\cite{Gong_eccv2018_pgn}.} CIHP is the most popular benchmark for multiple human parsing. As shown in Table~\ref{tab:sota}, QANet with ResNet50 is better than the current best methods~\cite{Yang_eccv2020_rprcnn, Ji_eccv2020_sematree} in terms of mIoU, AP$^\text{p}$ and AP$^\text{r}$. In terms of AP$^\text{r}$, QANet outperforms RP R-CNN~\cite{Yang_eccv2020_rprcnn} by 13.9 points, outperforms SemaTree~\cite{Yang_eccv2020_rprcnn} by 12.2 points. Such a performance gain is particularly impressive considering that improvement on multiple human parsing task is very challenging. Combined with the larger capacity backbone HRNet-W48, QANet achieves state-of-the-art on CIHP \texttt{val} set, which is 66.1\% mIoU, 64.5\% AP$^\text{p}$, 75.7\% PCP$_\text{50}$ and 60.8\% AP$^\text{r}$.

\vspace{6pt}
\noindent\textbf{MHP-v2~\cite{Zhao_mm2018_mhpv2}.} In the middle part of Table~\ref{tab:sota} also reports QANet performance on MHP-v2 \texttt{val} set. Considering that the previous methods do not provide AP$^\text{r}$ criterions, we only compare mIoU and AP$^\text{p}$. QANet with ResNet50 outperforms RP R-CNN~\cite{Yang_eccv2020_rprcnn} by 3.8 points in mIoU. With ResNet101, QANet outperforms CE2P~\cite{Ruan_aaai2019_ce2p} by 2.0 points in mIoU and 6.5 points in AP$^\text{p}$. 

\vspace{6pt}
\noindent\textbf{PASCAL-Person-Part~\cite{Xia_cvpr2017_ppp}.} In the lower part of Table~\ref{tab:sota}, we compare our method against recent methods on PASCAL-Person-Part \texttt{test} set. In term AP$^\text{p}$ and AP$^\text{r}$, our improvement is remarkable. QANet surpasses PGN~\cite{Gong_eccv2018_pgn}, Parsing R-CNN~\cite{Yang_cvpr2019_parsingrcnn} and RP R-CNN~\cite{Yang_eccv2020_rprcnn} more than 10.0 points in AP$^\text{p}$ and 15.0 points in AP$^\text{r}$.

\begin{figure*}
\begin{center}
\includegraphics[width=0.84\linewidth]{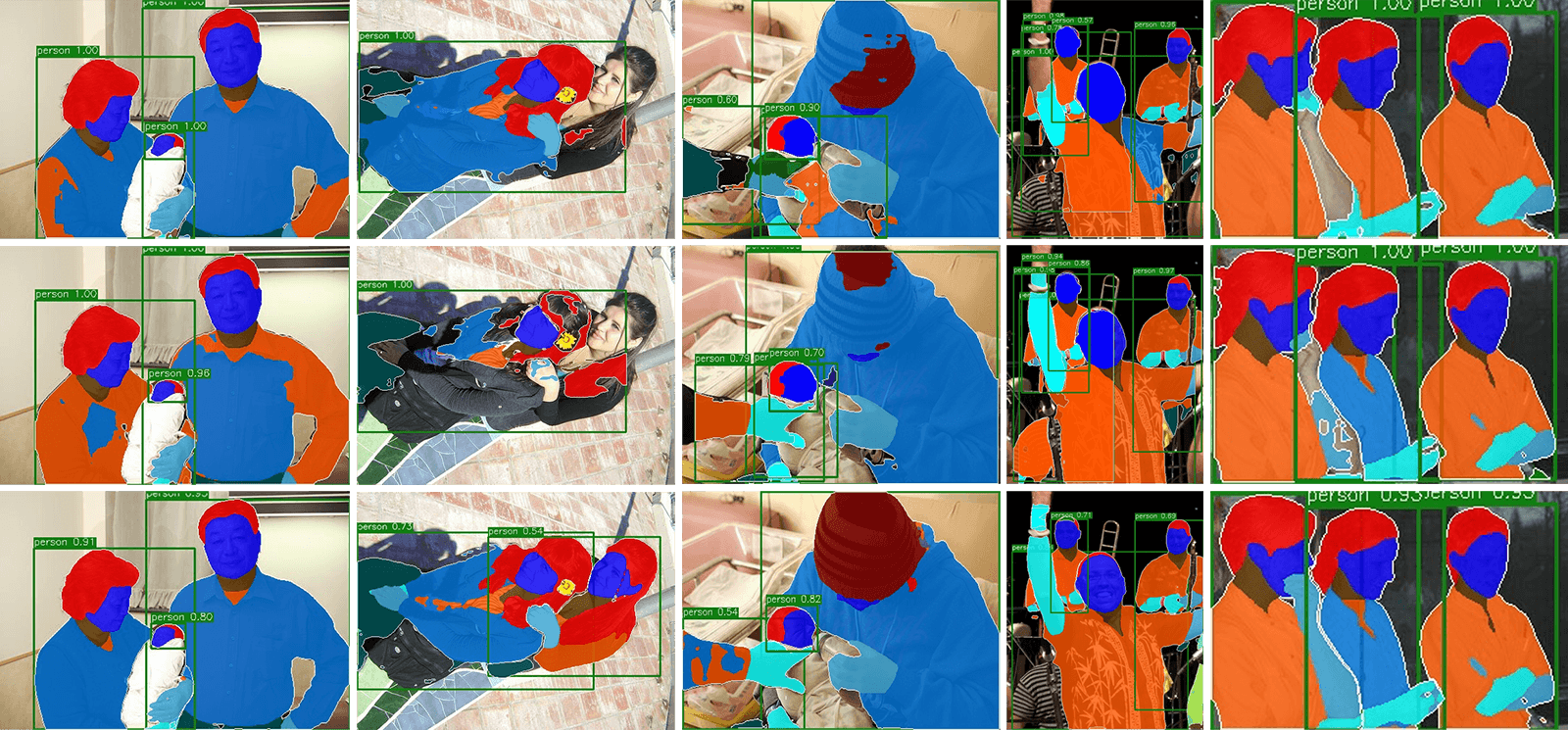}
\end{center}
\vspace{-1mm}
\caption{\textbf{Visual comparison on CIHP \texttt{val} set}. The images in the first row are the results of Parsing R-CNN~\cite{Yang_cvpr2019_parsingrcnn}, the second row are the results of RP R-CNN~\cite{Yang_eccv2020_rprcnn}, and the third row are the results of QANet. All methods are based on ResNet50.}
\label{fig:vis}
\vspace{-.8em}
\end{figure*}

\vspace{6pt}
\noindent\textbf{Qualitative Results.} Some qualitative comparison results on CIHP \texttt{val} set are depicted in Figure~\ref{fig:vis}. Qualitative comparison mainly shows the semantic segmentation ability of different methods. Compared with Parsing R-CNN~\cite{Yang_cvpr2019_parsingrcnn} (first row) and RP R-CNN~\cite{Yang_eccv2020_rprcnn} (second row), QANet (third row) has better performance in occlusion, human contour, complex background and confusing categories.

\subsection{Extensibility} 

\noindent\textbf{Region-based Human Parsing.} QAM can be regarded as a post-processing module, so it can be easily combined with other studies. Table~\ref{tab:region-based-parsing} shows the performance of QAM combined with two region-based multiple human parsing methods on CIHP \texttt{val} set. QAM improves Parsing R-CNN~\cite{Yang_cvpr2019_parsingrcnn} about 2.9 points in AP$^\text{p}$ and 10.7 points in AP$^\text{r}$. The IoU score has been used in RP R-CNN~\cite{Yang_eccv2020_rprcnn}, so QAM can improves 6.0 AP$^\text{r}$ and achieve similar AP$^\text{p}$. This is consistent with the experimental results in Row (d) and (e) of Table~\ref{tab:ablation_qweights}.

\begin{table}[t]
\centering
\small
\tabcolsep 0.04in 
\scalebox{0.82}{
\begin{tabular}{c|c|cc|cc}
  Methods                                                                                                &           QAM          &   AP$^\text{p}$        & AP$^\text{p}_\text{50}$  &  AP$^\text{r}$          & AP$^\text{r}_\text{50}$      \\
  \shline
  \multirow{2}{*}{Parsing R-CNN~\cite{Yang_cvpr2019_parsingrcnn}}   &                             & 53.9                         & 63.7                                & 36.5                         & 40.9                                   \\
                                                                                                                &      \checkmark     & \cgreen{55.8}{2.9}  & \cgreen{67.7}{4.0}           & \cgreen{47.2}{10.7} & \cgreen{53.8}{12.9}             \\
  \hline
  \multirow{2}{*}{RP R-CNN~\cite{Yang_eccv2020_rprcnn}}                  &                             & 59.5                         & 74.1                                 & 42.3                         & 48.2                                  \\
                                                                                                               &      \checkmark     & \cgreen{59.5}{0.0}   & \cgreen{74.2}{0.1}             & \cgreen{48.3}{6.0}    & \cgreen{55.5}{7.3}             \\
\end{tabular}
}
\vspace{.2em}
  \caption{\textbf{Comparison of region-based multiple human parsing on CIHP \texttt{val} set with or without QAM}. The quality weights of Parsing R-CNN~\cite{Yang_cvpr2019_parsingrcnn} is (1.0, \ \demph{0.0}, \ 1.0), and of RP R-CNN~\cite{Yang_eccv2020_rprcnn} is (3.0, \ 1.0, \ 1.0).}
  \label{tab:region-based-parsing}
\vspace{-.8em}
\end{table}

\vspace{6pt}
\noindent\textbf{Instance Segmentation.} Instance segmentation and multiple human parsing are highly similar in processing flow. Therefore, we believe that QAM (especially pixel score) is beneficial to the instance segmentation task. Table~\ref{tab:instance-seg} shows that results of some methods on COCO~\cite{Lin_eccv2014_coco} and LVISv1.0~\cite{Gupta_cvpr2019_lvis} with or without QAM\footnote{\fontsize{7pt}{1em}All experiments are based on \texttt{Detectron2}~\cite{Wu_url2019_detectron2}.}. On COCO \texttt{val} set, Mask R-CNN~\cite{He_iccv2017_maskrcnn} with QAM is better than the original counterpart: up to +1.0 AP$^\text{m}$. QAM can also slightly improve SOLOv2~\cite{Wang_nips2020_solov2}. On LVISv1.0 \texttt{val} set, we can also observe steady improvements.

\begin{table}[t]
\centering
\small
\tabcolsep 0.04in 
\scalebox{0.88}{
\begin{tabular}{c|c|c|ccc}
  Dataset                             & Methods                                                                                                &           QAM          &   AP$^\text{m}$        & AP$^\text{m}_\text{50}$  & AP$^\text{m}_\text{75}$      \\
  \shline
  \multirow{4}{*}{COCO}      & \multirow{2}{*}{Mask R-CNN~\cite{He_iccv2017_maskrcnn}}              &                             & 35.2                         & 56.2                                & 37.5                                           \\
                                            &                                                                                                               &      \checkmark     & \cgreen{36.2}{1.0}  & \cgray{55.5}{0.7}           & \cgreen{39.0}{1.5}                 \\
                                             \cline{2-6}
                                            & \multirow{2}{*}{SOLOv2~\cite{Wang_nips2020_solov2}}                      &                             & 34.8                         & 54.0                                & 37.1                                         \\
                                            &                                                                                                               &      \checkmark     & \cgreen{35.2}{0.4}  & \cgray{53.5}{0.5}           & \cgreen{37.7}{0.6}                  \\                                                                                                                                                             
  \hline
  \multirow{2}{*}{LVISv1.0}   & \multirow{2}{*}{Mask R-CNN~\cite{He_iccv2017_maskrcnn}}              &                             & 19.1                         & 30.2                                & 20.3                                         \\
                                            &                                                                                                               &      \checkmark     & \cgreen{20.1}{1.0}  & \cgreen{31.2}{1.0}           & \cgreen{21.4}{1.1}                 \\
\end{tabular}
}
\vspace{.2em}
  \caption{\textbf{Comparison of instance segmentation on COCO / LVISv1.0 \texttt{val} sets with or without QAM}. The quality weights is (1.0, \ \demph{0.0}, \ 1.0).}
  \label{tab:instance-seg}
\vspace{-.8em}
\end{table}

\begin{table}[t]\centering
\subfloat[\textbf{Track 1: Multi-Person Human Parsing Challenge}.
\label{tab:lip:track-1:sota}]{
\scalebox{0.84}{
\begin{tabular}{c|ccc|c}
  Methods                                                                                                 &        mIoU      &  AP$^\text{r}$ &      Rank   &                                   \\
  \shline
BJTU\_UIUC~\cite{Ruan_aaai2019_ce2p}                                              &       63.77        &        45.31      &      54.54   &   \demph{2018 1st}  \\
PAT\_CV\_HUMAN                                                                                   &       67.83        &        55.43      &      61.63   &                                 \\
BAIDU-UTS~\cite{Li_arxiv2019_schp}                                                     &       \textbf{69.10}        &        55.01      &      62.05   &   \demph{2019 1st}  \\
tfzhou.ai                                                                                                    &       68.69        &        56.40      &      62.55   &                                 \\
  \hline
\textbf{QANet (ours)}                                                                                &       68.76        &        \textbf{64.63}      &      \textbf{66.69}   &    \demph{2020 1st}  \\
\end{tabular}}}

\subfloat[\textbf{Track 2: Video Multi-Person Human Parsing Challenge}.
\label{tab:lip:track-2:sota}]{
\scalebox{0.84}{
\begin{tabular}{c|cccc|c}
  Methods                                                                                                 &       mIoU    &  AP$^\text{h}$ &  AP$^\text{r}$ &  Rank   &   \\
  \shline
BAIDU-UTS~\cite{Li_arxiv2019_schp}                                                     &       60.14    &        76.52      &      52.97         &    63.21 & \demph{2019 1st}  \\
keetysky                                                                                                    &       60.17    &        84.71      &      55.19         &    66.69 &                               \\
PAT\_CV\_HUMAN                                                                                   &       60.58    &        84.93      &      55.56         &    67.02 &                               \\
  \hline
\textbf{QANet (ours)}                                                                                &       \textbf{62.67}     &        \textbf{89.59}      &      \textbf{59.34}         &    \textbf{70.53} &  \demph{2020 1st}  \\
\end{tabular}}}\\\vspace{1mm}
\caption{\textbf{QANet wins 1st place in Track 1 \& 2 of CVPR2020 LIP Challenge}. Our results ensemble four QANet models.}
\label{tab:lip:lip2020}
\vspace{-.8em}
\end{table}

\vspace{6pt}
\noindent\textbf{Challenges.} Benefiting from the superiority of QANet, we win 1st place in Track 1 \& 2 of CVPR2020 LIP Challenge (see Table~\ref{tab:lip:lip2020}) with better human detector~\cite{Tian_iccv2019_fcos, Zhang_arxiv2020_resnest} and test-time techniques. By replacing parsing losses with dense pose losses, QANet achieves the same impressive result in dense pose estimation. In CVPR2019 COCO DensePose Challenge, QANet wins 1st place and was significantly ahead of the other competitors in all five metrics (see Table~\ref{tab:coco2019}).


\begin{table}[t]
\centering
\small
\tabcolsep 0.04in 
\scalebox{0.92}{
\begin{tabular}{c|ccc|cc|c}
  Methods                                                                                                 &       AP$^\text{dp}$     & AP$^\text{dp}_\text{50}$ & AP$^\text{dp}_\text{75}$ & AP$^\text{dp}_\text{M}$ & AP$^\text{dp}_\text{L}$   &   \\
  \shline
DensePose R-CNN~\cite{Guler_cvpr2018_densepose}                         &                58                &             90                       &                   70                  &                  51                  &                 61                     \\
Parsing R-CNN~\cite{Yang_cvpr2019_parsingrcnn}                               &                65                &             93                       &                   78                  &                  57                  &                 68                   &  \demph{2018 1st}      \\
Detectron2~\cite{Wu_url2019_detectron2}                                             &                67                &             92                       &                   80                  &                  60                  &                 70                      \\
  \hline
\textbf{QANet (ours)}                                                                               &                \textbf{72}                &            \textbf{94}                      &                   \textbf{85}                  &                  \textbf{65}                  &                 \textbf{74}                    &    \demph{2019 1st}        \\	
\end{tabular}
}
\vspace{.2em}
  \caption{\textbf{QANet wins 1st place in CVPR2019 COCO DensePose Challenge}. Our result is obtained without ensembling.}
  \label{tab:coco2019}
\vspace{-.8em}
\end{table}

\section{Conclusion} 
\label{sec:conc}
The quality estimation of human parsing is an important but long neglected issue. In this work, we propose a network that can generate high-quality parsing results and a simple, low-cost quality estimation method, named QANet. Its core components are pixel score based on probability map and the quality-aware module (QAM) which integrates different quality information. QANet has verified its effectiveness and advancement on several challenging benchmarks, and wins 1st in three popular challenges, including CVPR2020 LIP Track 1 \& Track 2 and CVPR2019 COCO DensePose. Beyond the human parsing task, it is possible to adopt QANet and QAM for other instance-level tasks, like instance segmentation and dense pose estimation. We hope to promote a new breakthrough in the form of quality estimation and help more visual tasks.

{\small
\renewcommand\UrlFont{\color{Gray}\ttfamily}
\bibliographystyle{ieee_fullname}
\bibliography{egbib}
}

\end{document}